\documentclass[conference]{IEEEtran}
\IEEEoverridecommandlockouts
\usepackage{multirow}
\usepackage{cite}
\usepackage{amsmath,amssymb,amsfonts}
\usepackage{algorithmic}
\usepackage{graphicx}
\usepackage{textcomp}
\usepackage{xcolor}
\usepackage{xurl}
\usepackage[colorlinks=true, breaklinks]{hyperref}
\def\BibTeX{{\rm B\kern-.05em{\sc i\kern-.025em b}\kern-.08em
    T\kern-.1667em\lower.7ex\hbox{E}\kern-.125emX}}
\begin{document}

\title{A Domain-Adapted Lightweight Ensemble for Resource-Efficient Few-Shot Plant Disease Classification\\
}

\author{

\IEEEauthorblockN{Anika Islam}
\IEEEauthorblockA{\textit{Department of Computer Science and Engineering} \\
\textit{Islamic University of Technology}\\
Gazipur, Bangladesh \\
anikaislam@iut-dhaka.edu}
\and
\IEEEauthorblockN{Tasfia Tahsin}
\IEEEauthorblockA{\textit{Department of Computer Science and Engineering} \\
\textit{Islamic University of Technology}\\
Gazipur, Bangladesh \\
tasfiatahsin@iut-dhaka.edu}
\and
\IEEEauthorblockN{Zaarin Anjum}
\IEEEauthorblockA{\textit{Department of Computer Science and Engineering} \\
\textit{Islamic University of Technology}\\
Gazipur, Bangladesh \\
zaarinanjum@iut-dhaka.edu}
\and
\IEEEauthorblockN{Md. Bakhtiar Hasan}
\IEEEauthorblockA{\textit{Assistant Professor}\\
\textit{Department of Computer Science and Engineering} \\
\textit{Islamic University of Technology}\\
Gazipur, Bangladesh \\
bakhtiarhasan@iut-dhaka.edu}
\and
\IEEEauthorblockN{Dr. Md. Hasanul Kabir}
\IEEEauthorblockA{\textit{Professor and Head of the Department}\\
\textit{Department of Computer Science and Engineering} \\
\textit{Islamic University of Technology}\\
Gazipur, Bangladesh \\
hasanul@iut-dhaka.edu}

}

\maketitle

\begin{abstract}
Accurate and timely identification of plant leaf diseases is essential for resilient and sustainable agriculture, yet most deep learning approaches rely on large annotated datasets and computationally intensive models that are unsuitable for data-scarce and resource-constrained environments. To address these challenges, we present a few-shot learning approach within a lightweight yet efficient framework that combines domain-adapted MobileNetV2 and MobileNetV3 models as feature extractors, along with a feature fusion technique to generate robust feature representation. For the classification task, the fused features are passed through a Bi-LSTM classifier enhanced with attention mechanisms to capture sequential dependencies and focus on the most relevant features, thereby achieving optimal classification performance even in complex, real-world environments with noisy or cluttered backgrounds.
The proposed framework was evaluated across multiple experimental setups, including both laboratory-controlled and field-captured datasets. On tomato leaf diseases from the PlantVillage dataset, it consistently improved performance across 1 to 15 shot scenarios, reaching \textbf{98.23~$\pm$~0.33\%} at 15 shot, closely approaching the \textbf{99.98\%} state-of-the-art benchmark achieved by a Transductive LSTM with attention, while remaining lightweight and mobile-friendly. Under real-world conditions using field images from the Dhan Shomadhan dataset, it maintained robust performance, reaching \textbf{69.28~$\pm$~1.49\%} at 15 shot and demonstrating strong resilience to complex backgrounds. Notably, it also outperformed the previous state-of-the-art accuracy of \textbf{96.0\%} on six diseases (apple, blueberry, and cherry) from PlantVillage, achieving \textbf{99.72\%} with only 15-shot learning.
With a compact model size of approximately 40 MB and inference complexity of $\approx 1.12$ GFLOPs, this work establishes a scalable, mobile-ready foundation for precise plant disease diagnostics in data-scarce regions.
\end{abstract}

\begin{IEEEkeywords}
Plant disease classification, Few-shot learning, Lightweight deep learning, MobileNet, Attention mechanism, Precision agriculture
\end{IEEEkeywords}
\section{Introduction}
Agriculture constitutes a fundamental pillar of the global economy, particularly in developing regions where it directly influences food security and socio-economic stability. Among the numerous challenges confronting this sector, plant leaf diseases remain a persistent threat, capable of reducing crop yields by up to 62\% under severe outbreaks \cite{parvin2024prevention}. Although countries with agriculture-driven economies are especially susceptible to these impacts, the problem is universal and affects crop production across diverse climatic and geographic contexts. Consequently, developing automated and efficient plant disease detection frameworks holds significant importance worldwide, as such systems can assist farmers in early diagnosis, mitigate yield losses, and foster sustainable agricultural practices.

Recent developments in deep learning have greatly advanced automatic plant leaf disease recognition, leveraging architectures such as convolutional neural networks (CNNs), vision transformers (ViTs), generative adversarial networks (GANs), and hybrid attention–convolution models. Systematic reviews highlight that between 2020 and 2025, over 278 studies investigated the classification, detection, and segmentation of leaf diseases using deep learning techniques \cite{Upadhyay2025_DeepLearningPlantDisease}. For instance, a hybrid framework coupling CNNs and ViTs achieved accuracies of up to 99.24 \% \&  98\% on apple and corn leaf disease datasets \cite{Aboelenin2025_HybridPlantLeafDisease}. A recent lightweight hybrid model (HPDC-Net) achieves over 99\% accuracy with only ~0.52M parameters and 0.06 GFLOPs, enabling real-time use on resource-limited devices \cite{Asghar2025_HPDCNet}. However, it was tested on laboratory-controlled images and lacks validation in real-world agricultural fields, highlighting a key gap. More broadly, a comprehensive review published in 2025 reported that CNN models such as VGG, ResNet, and EfficientNet continue to dominate in terms of performance, but also emphasized ongoing challenges in dataset diversity and model generalization \cite{Salka2025_PlantLeafDiseaseCNNReview,Elfouly2025_LargeScalePlantDisease}. In parallel, recent works have focused on limited-data regimes, for example, a few-shot classification framework for leaf diseases achieved 98.09 \% accuracy using only 80 training examples per class \cite{Ahmed2025_DExNetLeafDisease}. Despite these impressive results, the success of these deep learning approaches remains heavily dependent on large, well-labeled image datasets and substantial computational resources constraints that hinder deployment in many real world agricultural environments.

Few-shot learning (FSL) has emerged as a promising direction to address these limitations. By enabling models to learn and generalize effectively from only a few labeled samples per class, FSL reduces dependence on extensive datasets and mitigates overfitting in data-scarce conditions\cite{yang2022survey, Sun2023FewShotPlantDiseaseReview}. This capability is particularly valuable in agricultural applications, where labeled images of rare or newly emerging diseases are limited and collecting large datasets is time-consuming and costly. Furthermore, FSL has demonstrated strong potential for accurately identifying small and complex visual patterns found in leaves, fruits, weeds, and pests, even under variable environmental conditions \cite{ragu2023object}.

To ensure that such models are also practical for real-world use, computational efficiency becomes a critical consideration. Rather than relying on large-scale or computationally intensive networks, lightweight and optimized architectures are needed to maintain a balance between recognition performance and resource usage specially in field or resource‑constrained agricultural environments \cite{Wei2025LightweightFewShotCropPestDisease}.

Motivated by these challenges, this research proposes an efficient few-shot learning framework for plant leaf disease classification. The proposed approach integrates domain-adapted lightweight feature extractors (MobileNetV2 and MobileNetV3 variants) with a Bidirectional Long Short-Term Memory (Bi-LSTM) classifier enhanced by an attention mechanism. The MobileNet variants produce compact yet discriminative feature representations, while the Bi-LSTM captures inter-feature dependencies and the attention layer emphasizes disease-relevant patterns. Together, these components form a robust, data-efficient, and computationally lightweight framework suitable for real-world agricultural environments.
 
The key contributions of this research are summarized as follows:

\begin{itemize}
    \item \textbf{Addressing data scarcity through few-shot learning:} 
    The proposed framework demonstrates that few-shot learning (FSL) can effectively classify plant leaf diseases using a minimal number of labeled samples per class, significantly reducing dependence on large annotated datasets.

    \item \textbf{Design of a lightweight and efficient architecture:} 
    An optimized combination of MobileNetV2, MobileNetV3-Small, and MobileNetV3-Large was developed as domain-adapted feature extractors, integrated with a Bi-LSTM classifier enhanced by an attention mechanism. This configuration achieves high accuracy while maintaining low computational cost.

    \item \textbf{Comprehensive evaluation under controlled and real-world conditions:} 
    The framework was rigorously evaluated using both the \textit{PlantVillage} dataset (laboratory background) and the \textit{Dhan Shomadhan} dataset (field background). Experimental results show consistent improvements in accuracy across 1--15-shot scenarios, demonstrating strong generalization in diverse imaging conditions.

    \item \textbf{Suitability for mobile and edge deployment:} 
    Due to its compact model size ($\approx$40~MB) and low computational complexity ($\approx$1.12~GFLOPs), the proposed framework is well-suited for deployment on mobile and other resource-constrained devices, enabling practical field-level plant disease diagnostics.
\end{itemize}

\section{Related Works}
Deep learning has significantly advanced plant leaf disease classification by leveraging convolutional neural networks (CNNs) for automated feature extraction and high-accuracy prediction. Studies such as Chelladurai \textit{et al}. \cite{chelladurai2025classification} combined U-Net for segmentation with VGG-16 and a transductive LSTM with attention, achieving 99.98\% accuracy on tomato leaf diseases. Similarly, Joseph \textit{et al.} \cite{joseph2024realtime} developed new datasets for rice, wheat, and maize covering major fungal and bacterial diseases, achieving testing accuracies up to 97.28\% (rice), 96.32\% (wheat), and 95.80\% (maize) by experimenting with eight fine-tuned CNNs such as Xception and MobileNetV2. Their proposed custom CNN trained from scratch also performed competitively, with accuracies of 98.08\% (wheat), 97.06\% (rice), and 97.04\% (maize). While these approaches show impressive results, they typically rely on large volumes of labeled data and involve computationally intensive models, making them less suitable for scenarios with limited data or low-resource deployment environments.

Few-shot learning has gained significant attention for plant disease classification due to its ability to address data scarcity by enabling models to learn new categories from limited labeled samples, thereby improving efficiency and reducing overfitting by utilizing techniques including transfer learning, meta-learning, data augmentation, and multimodal learning \cite{liu2022few}.

Meta-learning, or "learning to learn," facilitates rapid adaptation to new tasks through prior experiences with similar tasks, with main approaches including metric-based, model-based, and optimization-based methods \cite{gharoun2023metalearning, wu2023metalearning}. 
From various studies, metric-learning architectures such as Siamese Network \cite{koch2015siamese}, Prototype Network \cite{snell2017prototypical}, and Matching Network \cite{vinyals2016matching} have outperformed traditional CNNs when working with limited data. Wang \textit{et al}. \cite{wang2019plant} introduced a Siamese network-based few-shot learning approach for leaf classification, using a parallel two-way CNN with shared weights to learn a metric space optimized for proximity between similar samples and separation of dissimilar ones. This metric space was further refined using a spatial structure optimizer (SSO), and a k-nearest neighbor (kNN) classifier was applied for accuracy evaluation on small supervised datasets (Flavia, Swedish, Leafsnap). Li \textit{et al}. \cite{li2021metalearning} developed a task-driven meta-learning approach for pest and plant recognition, analyzing single-domain and cross-domain scenarios and evaluating accuracy across 36 experiments to establish baselines and investigate the effects of N-way, K-shot, and domain shift on classification. Mu \textit{et al}. \cite{mu2024contrastive} presented a two-phase algorithm combining supervised contrastive learning and meta-learning, where an encoder is pre-trained with supervised contrastive loss and used with a nearest-centroid classifier, achieving 79.51\% accuracy on potato leaf disease detection with only 30 training images and reduced GPU requirements using small batch sizes. Additionally, Uzhinskiy \cite{uzhinskiy2025evaluation} systematically evaluated various metric-learning losses on a 68-class plant disease dataset (4000 images), finding that angular/cosine-margin methods like CosFace and ArcFace outperformed traditional Siamese and triplet losses, offering effective strategies for few-shot plant disease recognition under limited data. Despite these advancements, Wang \textit{et al}. \cite{wang2019plant} noted that high inter-class variation in most datasets can oversimplify classification; meta-learning models also present deployment challenges due to high storage and computation demands, limiting their feasibility for edge devices and thus hindering broader agricultural adoption \cite{wu2023metalearning}. Furthermore, while traditional models perform well on common disease, they struggle with complex or multi-disease cases, especially under limited data, making accurate identification of specific bacterial clusters difficult \cite{wu2023metalearning}.

Another machine learning technique is transfer learning where a model trained on one task is re-purposed for a second, related task \cite{machinelearningmastery_transfer_learning}. Essentially, the goal is to apply the model's learnings from a task with a large amount of labeled training data to a new task with less data. This addresses the challenges faced with mneta learning. Instead of starting the learning process from scratch, the model leverages patterns acquired from resolving a related task. Reusing or transferring information from previously learned tasks to new tasks has the potential to significantly improve learning efficiency \cite{george2019self}.\\
In our research work, we have explored the pre-trained model approach to transfer learning. The approach includes three prominent steps - i) Select Source Model, ii) Reuse Model and iii) Tune Model. Transfer learning-based techniques with pre-trained architectures can be applied to generate highly general feature representations for FSL, according to recent works. Chen \textit{et al} propose a simple method that uses a single binary classifier with class identifiers as input prefixes to learn all few-shot classes jointly \cite{chen2019closer}.The paper also shows that this method outperforms most existing meta-learning methods on a standard few-shot text classification dataset. According to Tian \textit{et al}, transfer learning learns better embeddings than traditional meta-learning approaches \cite{tian2020rethinking}. The paper argues that the key factor for few-shot learning is the quality of the feature embeddings rather than the meta-learning algorithm. By learning a good embedding model on a large-scale classification task, transfer learning can achieve better generalization than meta-learning methods that learn embeddings on smaller and more diverse tasks. Another study \cite{dvornik2020selecting} proposes a method for few-shot classification that uses feature selection from a multi-domain representation. The main idea is to leverage a set of feature extractors that are trained on different visual domains and then select the most relevant features for a new task which is a  simpler and more effective approach than feature adaptation, that is a common technique for transfer learning. The study demonstrates experimentally that a library of pre-trained feature extractors combined with a simple feed-forward network learned with an L2-regularizer can be an excellent option for solving cross-domain few-shot image classification. Their experimental results suggest that this simpler sample-efficient approach far outperforms several well-established meta-learning algorithms on a variety of few-shot tasks.\\
\section{Materials and methods}
\subsection{Dataset}
Our experiments utilized two datasets: the PlantVillage \cite{DBLP:journals/corr/HughesS15} dataset and the Dhan Shomadhan \cite{hossain2023dhan} dataset. The PlantVillage dataset, which contains leaf images captured against plain backgrounds, was used for framework development and efficiency testing. Its large-scale and class-balanced image set makes it well suited for few-shot learning, allowing models to be trained effectively with a small number of examples per class. This is particularly advantageous for plant disease classification, where labeled data is often scarce.

To evaluate the robustness of our proposed framework in real-world scenarios with complex backgrounds, we employed the Dhan Somadhan dataset, specifically utilizing the field background variation of diseased rice leaf images. This version of the dataset captures leaves in their natural agricultural settings with complex and diverse backgrounds, thereby reflecting practical field conditions more accurately and providing a meaningful benchmark for assessing the model’s generalization and effectiveness outside controlled environments.

The following subsections provide detailed descriptions of each dataset used in this study.
\subsubsection{PlantVillage}
The PlantVillage dataset comprises 54,303 images, featuring healthy and unhealthy leaves from 14 crop species with 26 diseases. It is categorized into 38 groups based on species and disease. Initially, the dataset was designed to support the development of mobile disease diagnostics using ML.

\begin{figure}[!htbp]
    \centering
    \includegraphics[width=\linewidth]{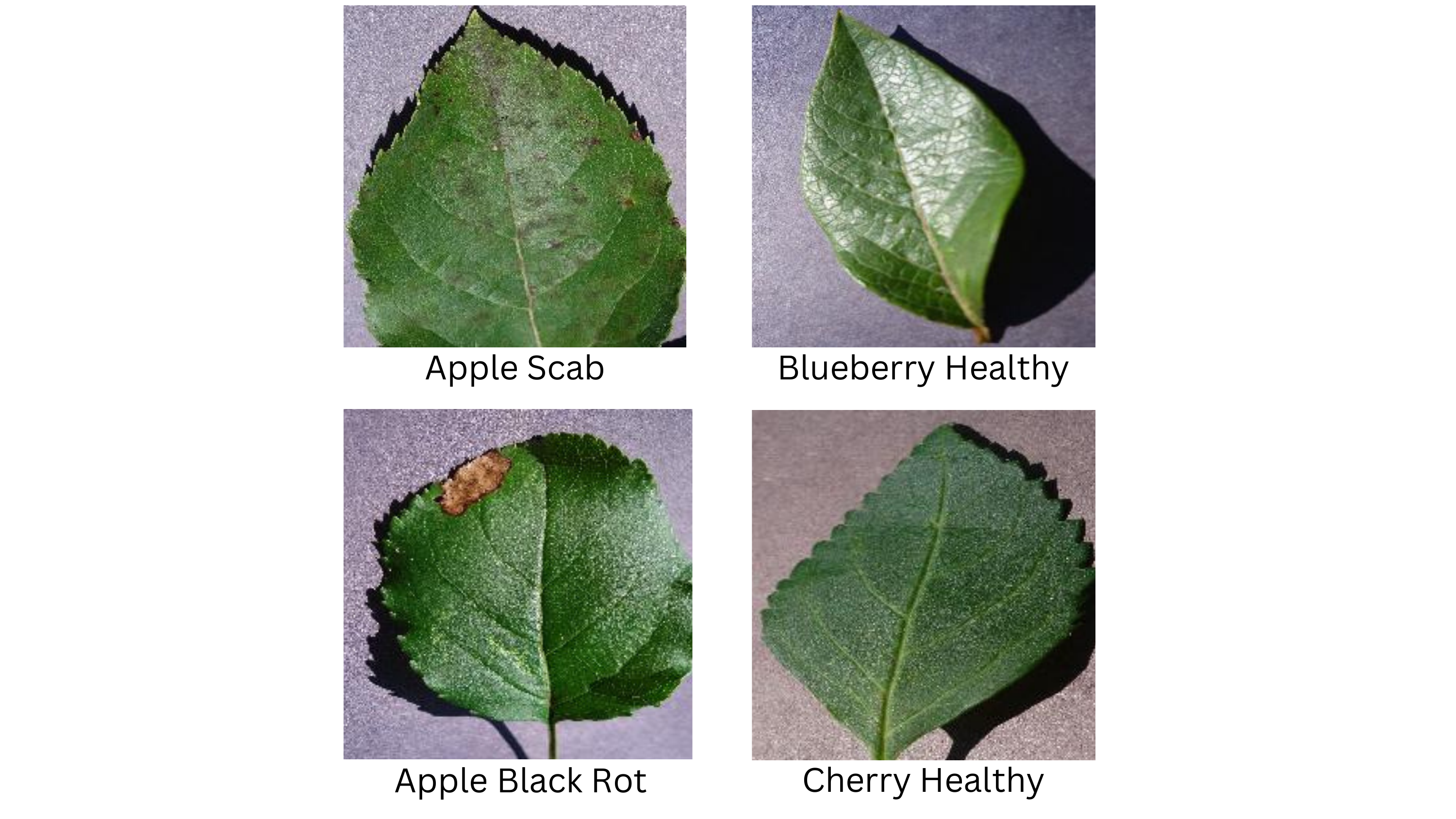}    
    \caption{Plant Village Dataset (Adapted from \cite{DBLP:journals/corr/HughesS15})}
    \label{fig:PlantVillage}
\end{figure}

The class names, along with the corresponding crop types and the number of images per class in the PlantVillage dataset, are summarized in Table \ref{plantVillageTable}, while Fig. \ref{fig:PlantVillage} illustrates a few sample images from the dataset.
\begin{table}[htbp]
\centering
\caption{Overview of Crops, Class Names, and Total Images in the PlantVillage Dataset}
\renewcommand{\arraystretch}{1.5} 
\setlength{\tabcolsep}{3pt} 
\resizebox{\linewidth}{!}{%
\begin{tabular}{|p{3cm}|p{6cm}|c|}
\hline
\textbf{Crop} & \textbf{Class Name} & \textbf{Number of Images} \\ \hline
Apple & Apple scab & 630 \\ \hline
Apple & Black rot & 621 \\ \hline
Apple & Cedar apple rust & 275 \\ \hline
Apple & healthy & 1645 \\ \hline
Blueberry & healthy & 700 \\ \hline
Cherry (including sour) & Powdery mildew & 441 \\ \hline
Cherry (including sour) & healthy & 854 \\ \hline
Corn (maize) & Cercospora leaf spot / Gray leaf spot & 1191 \\ \hline
Corn (maize) & Common rust & 1192 \\ \hline
Corn (maize) & Northern Leaf Blight & 985 \\ \hline
Corn (maize) & healthy & 1187 \\ \hline
Grape & Black rot & 1084 \\ \hline
Grape & Esca (Black Measles) & 696 \\ \hline
Grape & Leaf blight (Isariopsis Leaf Spot) & 1204 \\ \hline
Grape & healthy & 1082 \\ \hline
Orange & Haunglongbing (Citrus greening) & 550 \\ \hline
Peach & Bacterial spot & 265 \\ \hline
Peach & healthy & 492 \\ \hline
Pepper (bell) & Bacterial spot & 984 \\ \hline
Pepper (bell) & healthy & 1476 \\ \hline
Potato & Early blight & 1000 \\ \hline
Potato & Late blight & 1000 \\ \hline
Potato & healthy & 1528 \\ \hline
Raspberry & healthy & 613 \\ \hline
Soybean & healthy & 509 \\ \hline
Squash & Powdery mildew & 825 \\ \hline
Strawberry & Leaf scorch & 684 \\ \hline
Strawberry & healthy & 1117 \\ \hline
Tomato & Bacterial spot & 940 \\ \hline
Tomato & Early blight & 1000 \\ \hline
Tomato & Late blight & 1000 \\ \hline
Tomato & Leaf Mold & 952 \\ \hline
Tomato & Septoria leaf spot & 1771 \\ \hline
Tomato & Spider mites / Two-spotted spider mite & 1740 \\ \hline
Tomato & Target Spot & 1404 \\ \hline
Tomato & Tomato Yellow Leaf Curl Virus & 535 \\ \hline
Tomato & Tomato mosaic virus & 373 \\ \hline
Tomato & healthy & 1591 \\ \hline
\end{tabular}%
}
\label{plantVillageTable}
\end{table}

\subsubsection{Dhan Shomadhan Dataset}
Dhan-Shomadhan is a comprehensive dataset featuring 1,106 images of rice leaves affected by five different diseases: Brown Spot, Leaf Scaled, Rice Blast, Rice Turngo, and Steath Blight. These images are captured in two background variations, namely field background and white background. This dataset is designed to facilitate research in rice leaf disease classification and detection using computer vision and pattern recognition techniques.

\begin{figure}[!htbp]
    \centering
    \includegraphics[width=\linewidth]{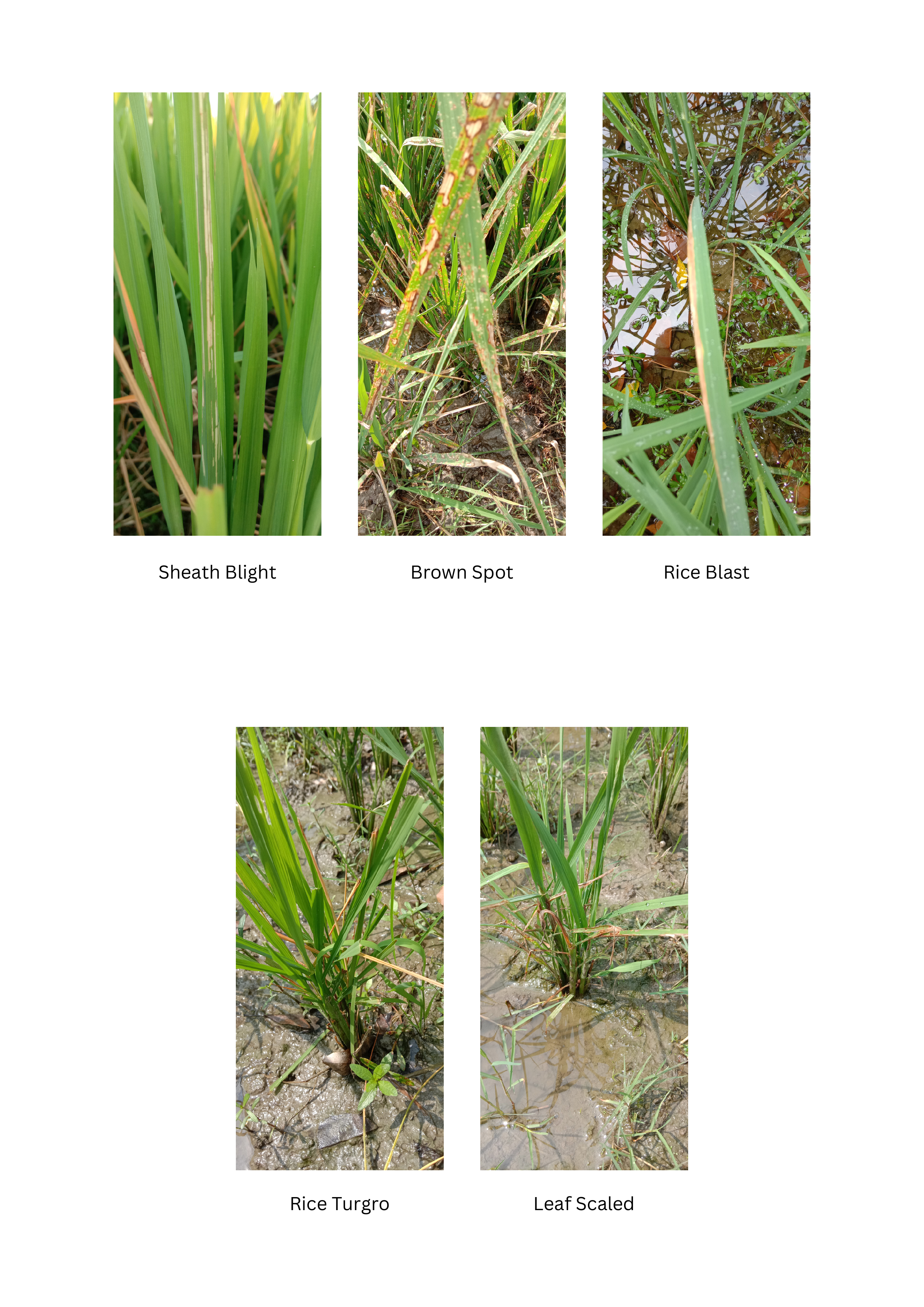}    
    \caption[Dhan Shomadhan Dataset]{Dhan Shomadhan Dataset (Adapted from \cite{hossain2023dhan})}
    \label{fig:Dhanshomadhan}
\end{figure}

A selection of sample images from the Dhan Shomadhan dataset, captured in field environments, is shown in Fig. \ref{fig:Dhanshomadhan}.
\subsection {Proposed Framework}
The goal of the proposed framework is to develop an efficient and robust few shot learning (FSL) model for plant leaf disease detection that can operate effectively in real world agricultural environments with complex backgrounds and minimal labeled training data. In practice, acquiring large scale annotated datasets of diseased plant leaves is time consuming, labor intensive, and often infeasible for newly emerging diseases. Our framework addresses this labeled data scarcity problem by leveraging FSL, enabling rapid adaptation to novel disease classes using only a few examples. Additionally, the framework is designed to be lightweight, computationally efficient, and mobile friendly, making it suitable for deployment on low power edge devices and agricultural robots operating in the field. This ensures that the solution remains practical for real time detection in resource constrained environments.

\begin{figure}[t]
    \centering
    \includegraphics[width=0.9\columnwidth]{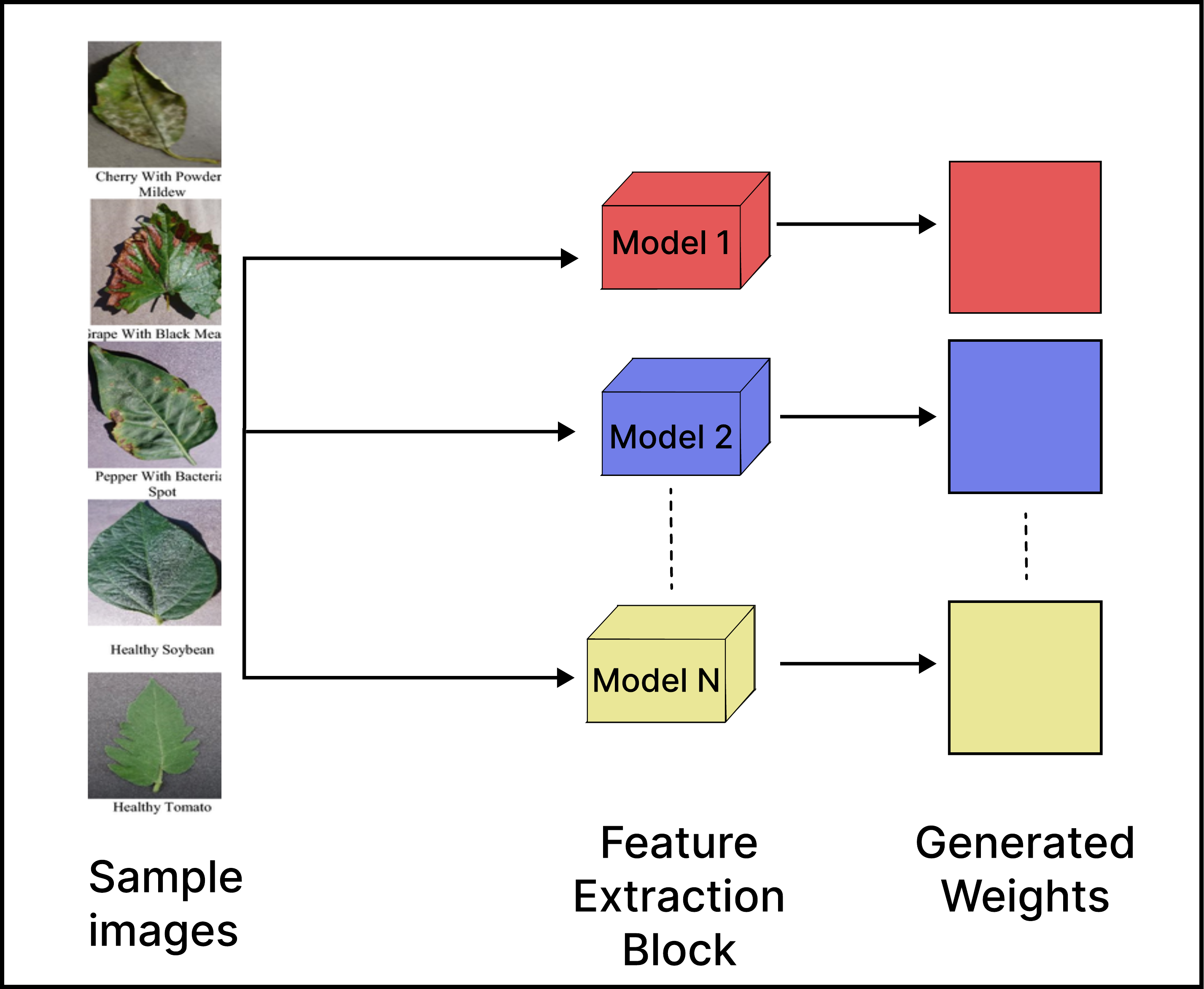}
    \caption{Fine Tuning}
    \label{fig:Frame_1}
\end{figure}

\begin{figure}[t]
    \centering
    \includegraphics[width=0.9\columnwidth]{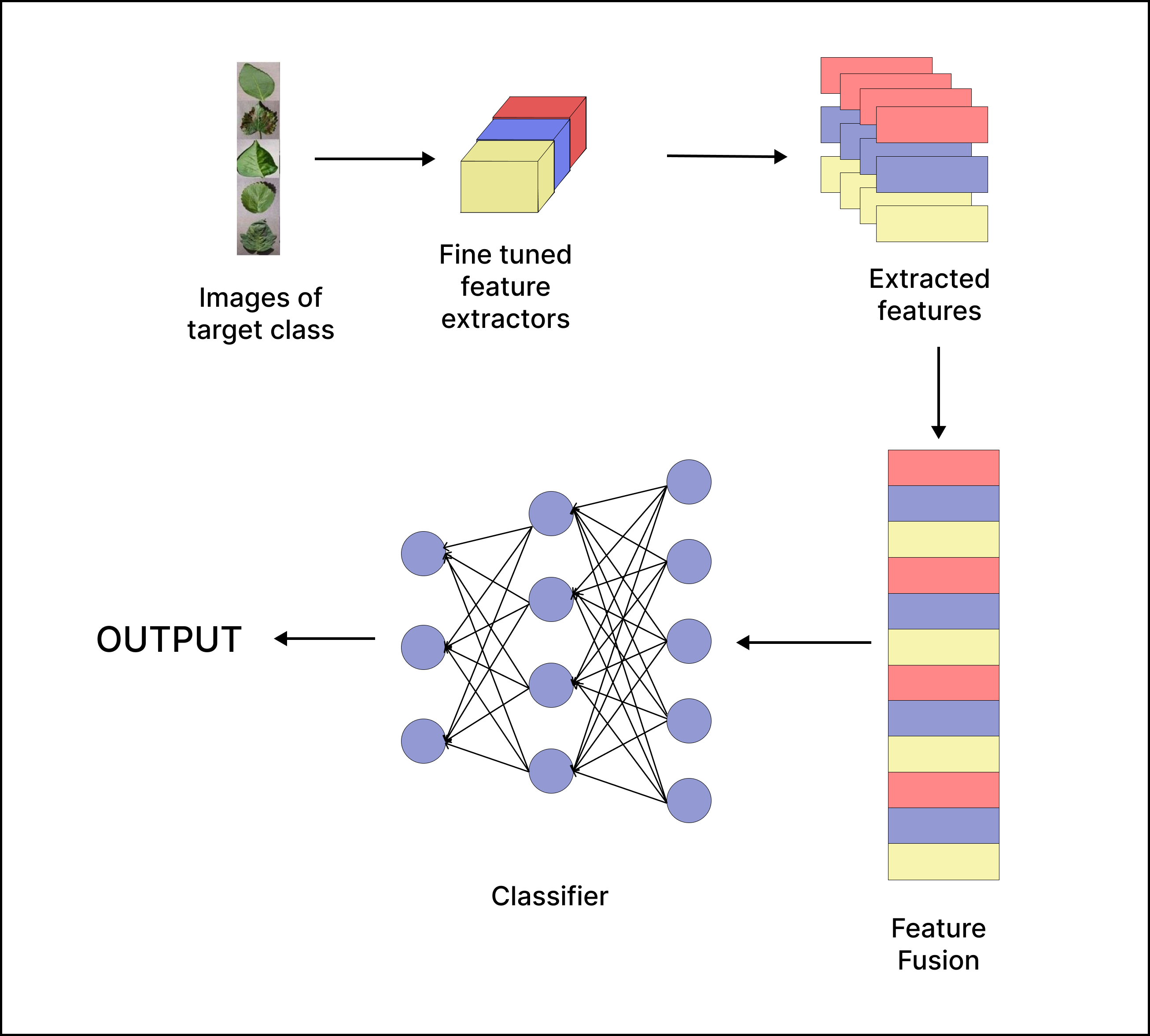}
    \caption{Feature Fusion and Classification}
    \label{fig:Frame_2}
\end{figure}

Our framework contains four main components: the Domain Adaptation Block (DAB), Feature Extraction Block (FEB), Feature Fusion Block (FFB), and Classification Block (CB). Three lightweight pre-trained CNN models, MobileNetV2, MobileNetV3 Small, and MobileNetV3 Large are employed as feature extractors. These models are first domain adapted using the Plant Village dataset to align their feature representations with the characteristics of plant leaf images. The outputs from the three domain-adapted models capture complementary visual cues due to differences in their architectural designs and receptive fields. These feature vectors are then concatenated to form a richer, multi-scale feature representation. This combined representation is subsequently processed by a Bidirectional Long Short-Term Memory (BiLSTM) network, which models inter-feature dependencies in both forward and backward directions, thereby enhancing discriminative capability. An attention mechanism is integrated on top of the BiLSTM to focus the model’s learning on the most informative features relevant to disease patterns, while suppressing irrelevant background noise. The final classification layer assigns the input leaf image to the appropriate disease category, in FSL scenarios.
\subsubsection{Feature Extractors}
While conventional deep learning methods require large-scale labeled datasets to perform effectively, such data demands are unrealistic in few-shot learning (FSL) scenarios. To address this limitation, we adopt a multi-expert feature extraction strategy using three complementary lightweight pretrained CNN architectures: MobileNetV2 \cite{sandler2018mobilenetv2}, MobileNetV3  Small and Large\cite{howard2019mobilenetv3}. This design is motivated by the principle that different architectures emphasize distinct aspects of feature representation; by exposing a few samples to multiple ``critics,'' each with its own inductive biases and receptive field characteristics, we obtain richer feature observations with minimal computational cost. 

To set a benchmark, we first experimented with pretrained MaxViT, ConvNeXt, and Swin Transformer models, selected for their high Top-1 and Top-5 ImageNet-1K accuracy and comparatively lower computational demands which outperformed the nine residual and dense CNN feature extractors of DExNet \cite{Ahmed2025_DExNetLeafDisease} while requiring only three models and about half the total computational resources. To further reduce cost, we evaluated smaller pretrained models and ultimately found that the MobileNetV2, MobileNetV3 Large, and MobileNetV3 Small combination not only lowered total model size and GFLOPs significantly but also matched or even exceeded our benchmark accuracy in FSL settings.

The superior performance of this trio can be attributed to their architectural efficiency and complementary design. MobileNetV2 employs an inverted residual structure with linear bottlenecks, which preserves representational capacity while minimizing memory access costs. MobileNetV3-Small and MobileNetV3-Large, in contrast, incorporate Squeeze-and-Excitation (SE) attention modules and optimized hard-swish activation functions to adaptively recalibrate feature channels, thereby improving the discrimination of fine-grained disease patterns. The Small variant is optimized for extreme resource constraints, ensuring robust performance even on mobile-grade processors, while the Large variant retains higher channel capacity for capturing complex textural and color variations. Collectively, these three models generate feature embeddings that are both computationally efficient and highly robust to variations in leaf textures, colors, and background noise.  
Compared to larger vision transformers or ensembles of deep residual networks, this three-expert setup achieves superior trade-offs between accuracy, memory footprint, and inference speed, making it particularly suitable for FSL-driven plant disease detection on mobile devices and agricultural robots.
\subsubsection{Domain Adaptation Block (DAB)}

The Domain Adaptation Block (DAB) is designed to reduce the gap between the 
generic pre-trained MobileNet models and the target domain of plant leaf images with diverse backgrounds, lighting variations, and noise. Since MobileNet variants are originally trained on large-scale datasets such as ImageNet, their representations may not capture the fine-grained textures, color variations, and disease symptoms present in agricultural images.

To address this, we fine-tuned each MobileNet variant using 28 classes from the PlantVillage dataset, excluding tomato leaf samples. The reason behind this choice is that, while the models are already capable of extracting generic features, they can be further specialized by learning from a large collection of diverse leaf samples. 
By excluding tomato leaves, we ensure that during adaptation the models never see the target domain data of our experiments.

After fine-tuning, the adapted models evolve into ``expert'' feature extractors. They learn to recognize what constitutes a healthy or diseased leaf, the discriminative 
features that are important (e.g., lesion texture, color distortions, venation patterns), and how to differentiate disease categories. Consequently, when only a few target 
samples are available in the few-shot learning phase, these expert models can extract richer and more discriminative features compared to their non-adapted counterparts, thereby improving downstream classification.
\subsubsection{Feature Extraction Block (FEB) and Feature Fusion Block (FFB)}

The Feature Extraction Block (FEB) is responsible for generating robust representations from input leaf images. Each domain-adapted MobileNet model processes the image independently, producing a high-dimensional feature vector that captures spatial and semantic information relevant to leaf health and disease patterns.Formally, if $f_{\theta_i}(x)$  represents the feature vector generated by the $i^{\text{th}}$ MobileNet variant for an input leaf image $x$, then the representation set of FEB can be defined as:

\[
\mathcal{F}(x) = \left\{ f_{\theta_1}(x), f_{\theta_2}(x), f_{\theta_3}(x) \right\},
\]

where $i=1,2,3$ corresponds to MobileNetV2, MobileNetV3-Small, and MobileNetV3-Large, respectively. Each $f_{\theta_i}(x)$ encodes different spatial and channel-wise characteristics due to architectural differences, producing complementary feature embeddings that highlight different aspects of the input, such as texture, color variation, and structural patterns.

The Feature Fusion Block (FFB) then merges the outputs from the three feature extractors to form a single, comprehensive feature representation. The fused feature vector $\mathcal{F}_{fusion}(x)$ can be defined as:

\[
\mathcal{F}_{fusion}(x) = \text{Concat} \Big( \mathcal{F}_1(x), \mathcal{F}_2(x), \mathcal{F}_3(x) \Big),
\]

where $\text{Concat}(\cdot)$ denotes the concatenation operation. By combining the embeddings, the FFB captures multi-scale and complementary information from each backbone, enhancing the overall discriminative capability. This fused representation serves as a unified input for the subsequent classification block, ensuring that the model can make accurate predictions even in few-shot scenarios, while maintaining robustness to variations in leaf appearance and background conditions.

\subsection{Classification Block (CB)}

The Classification Block (CB) is designed to transform the fused feature representation into a final disease prediction while maximizing discriminative power under few-shot conditions. After the Feature Fusion Block (FFB), the concatenated embeddings retain multi-scale and complementary cues, but they exist as a high-dimensional vector that may contain redundant or weakly relevant information. To refine this representation, we employ a Bidirectional Long Short-Term Memory (BiLSTM) network, which models inter-feature dependencies in both forward and backward directions. This allows the model to capture contextual relationships among feature dimensions, enabling it to learn richer patterns of variation in leaf textures and disease symptoms.

To further enhance feature discrimination, an attention mechanism is integrated on top of the BiLSTM outputs. The attention layer adaptively assigns higher weights to the most informative dimensions of the fused representation while suppressing irrelevant or noisy components, such as background textures or lighting artifacts. This selective focus ensures that the model emphasizes subtle yet crucial disease-related patterns that may otherwise be overlooked in few-shot scenarios.

Finally, the attended feature vector is passed through a fully connected layer followed by a softmax activation function to assign the input image to the appropriate disease class. By combining BiLSTM’s sequential modeling with the selective power of attention, the Classification Block ensures robust generalization even with very limited training samples, enabling accurate recognition of diverse plant leaf diseases in both controlled and field conditions.

\section{Result Analysis}

\subsection{Evaluation Metrics} 
In our study, we use several key metrics to evaluate the classification performance of plant leaf disease models. The primary metric is the \textbf{average accuracy} with a 95\% confidence interval (CI95). Additionally, we report \textbf{per-class precision}, \textbf{recall}, \textbf{F1-score}, and \textbf{support} to evaluate class-wise performance, which is critical in multiclass classification tasks with class imbalance:
\subsubsection{Accuracy}
Accuracy is the ratio of correctly predicted instances to the total instances.

\[
\text{Accuracy} = \frac{TP + TN}{TP + TN + FP + FN}
\]

where:
\begin{itemize}
  \item \( TP \) is True Positives,
  \item \( TN \) is True Negatives,
  \item \( FP \) is False Positives,
  \item \( FN \) is False Negatives.
\end{itemize}
\subsubsection{Average Accuracy and Confidence Interval}
It is calculated as the mean accuracy over multiple repetitions of the experiment. The 95\% confidence interval (CI95) provides a range around the average accuracy, signifying a 95\% confidence that the true accuracy of the model falls within this interval.

\[
\text{Average Accuracy} = \frac{1}{n} \sum_{i=1}^{n} \text{Accuracy}_i
\]

where \( n \) is the number of repetitions, and \( \text{Accuracy}_i \) is the accuracy of the \( i \)-th repetition.\\
For example, an average accuracy of 80\% ± 2\% means we are 95\% confident that the model's true accuracy lies between 78\% and 82\%.

Average accuracy was used as the primary evaluation metric to provide a robust and reliable estimate of overall model performance across multiple experimental runs, reducing the impact of random variation. This is particularly important in plant leaf disease classification tasks, where performance can fluctuate due to class imbalance and limited data. The inclusion of the 95\% confidence interval further strengthens the reliability of the results by quantifying the uncertainty around the average performance.

\subsection{Experimental Setup}
\subsubsection{Setup 1: Pipeline Development and Robustness Testing}
\begin{itemize}
    \item In the meta-training phase of this experiment, we fine-tuned our feature extractors using 28 classes from the PlantVillage dataset, excluding tomato leaf disease classes.
    \item For the meta-testing phase, we focused on 10 classes specifically from tomato leaf images.
    \item Within these classes, we divided the data into Support and Query sets at an 80:20 ratio.
    \item Each task involved randomly choosing \(K\) support samples and 50 query samples per class from the available images.
    \item To ensure unbiased results, we repeated this setup 100 times, with each repetition involving the random selection of \(K\) samples.
    \item The reported performance is the average over these 100 repetitions.
\end{itemize}
\subsubsection{Setup 2: Adaptation to Argueso \textit{et al}. \cite{argueso2020fewshot} Experimental Setup}
\begin{itemize}
    \item We adopted the experimental setup proposed by Argueso \textit{et al}. \cite{argueso2020fewshot}.
    \item In the meta-testing phase, we specifically selected 6 classes representing apple, blueberry, and cherry leaf diseases from the Plant Village Dataset.
    \item Notably, we avoided fine-tuning the pre-trained feature extractors since, in the first experiment, the pre-trained models had already seen images of those classes.
    \item The data set was divided into Support and Query sets in an 80:20 ratio.
    \item For each task, we randomly selected \(K\) support samples and 50 query samples per class from the available images.
    \item To ensure unbiased results, we repeated this setup 20 times, with each repetition involving the random selection of \(K\) samples.
    \item The reported results reflect the average performance obtained from these repetitions.
\end{itemize}
\subsubsection{Setup 3: Evaluation in Real-life Scenarios with complex backgrounds}
\begin{itemize}
\item In the meta-training phase of this experiment, we fine-tuned our feature extractors using 28 classes from the PlantVillage dataset, excluding tomato leaf disease classes.
    \item For the meta-testing phase, we focused on 5 classes specifically from Paddy leaf disease images from \cite{hossain2023dhan}.
    \item Within these classes, we divided the data into Support and Query sets at an 80:20 ratio.
    \item Each task involved randomly choosing \(K\) support samples and the entire test set as query samples per class from the available images.
    \item To ensure unbiased results, we repeated this setup 100 times, with each repetition involving the random selection of \(K\) samples.
    \item The reported performance is the average over these 100 repetitions.
\end{itemize}

\subsection{Experiment With Tomato Leaf Class }
\subsubsection{Performance Using Advanced Feature Extractors}
At first, we thoroughly analyzed the performance of the ensemble mentioned by Chowdhury \textit{et al.} \cite{chowdhury2021fewshot}, comprising nine feature extractors: ResNet18, ResNet34, ResNet50, ResNet101, ResNet152, DenseNet121, DenseNet161, DenseNet169, DenseNet201, as well as the ensembles of MaxViT, ConvNeXt, and Swin Transformer—which we selected for their high accuracy on ImageNet-1K—across various k-shot scenarios in Experimental Setup-1. We used Dense, LSTM, and Bi-LSTM as classifiers. Furthermore, we compared the total model size and total GFLOPs of these ensembles.

In the results Table \ref{Result1}, we also carefully analyzed and assessed our implemented MaxViT, ConvNeXt, and Swin Transformer feature extractor combination against the feature extractor from the paper \cite{chowdhury2021fewshot}, within the same experimental setup.

Regarding accuracy evaluations, we tested the performance of both frameworks in scenarios with 1, 5, 10, and 15 shots. For the classifier, we conducted experiments using a Dense classifier in the Chowdhury \textit{et al.} \cite{chowdhury2021fewshot} framework and Dense, LSTM, and Bi-LSTM in our framework with ensembles of MaxViT, ConvNeXt, and Swin Transformer as feature extractors. The findings in  Table \ref{Result1} consistently showed that, in all shot scenarios, our framework with Bi-LSTM outperformed all other combinations. 

Notably, our combination of Ensemble of  MaxViT, ConvNeXt, and Swin Transformer feature extractors showed approximately half the total model size and Gflops compared to the reference. That means we were able to optimize the framework's implementation and architecture by reducing the amount of storage space required and the computational power needed for inference or training through the reduction of the total model size and GFLOPs of the feature extractor combination, making it more computationally efficient than the baseline model of Chowdhury \textit{et al.} \cite{chowdhury2021fewshot}.

\subsubsection{Selection of Computationally Efficient Feature Extractors}
Although we have significantly reduced the requirements for storage and computational power, it wasn't sufficient. Our aim is to utilize this framework with embedded systems like mobile phones, for which we need to further reduce storage requirements and computational demands. The total model size and computational power requirement of the feature extractor ensemble comprising MaxViT, ConvNeXt, and Swin Transformer still remains very high at 900.01 MB and 47.01 GFLOPs, respectively. This makes it less efficient for incorporation into a mobile app.

\begin{table*}[htbp] 
    \centering
    \caption{Performance of Base Framework \cite{chowdhury2021fewshot} in Our Experimental Setup-1 Compared to Our Implementation With Advanced Feature Extractors}
    \renewcommand{\arraystretch}{1.5} 
    \setlength{\tabcolsep}{3pt} 
    \resizebox{\linewidth}{!}{ 
        \begin{tabular}{|p{5cm}|p{2cm}|c|c|c|c|c|c|}
            \hline
            \textbf{Feature Extractor} & \textbf{Total Model Size (MB)} & \textbf{GFLOPS} & \textbf{Classifier} & \textbf{1 Shot} & \textbf{5 Shot} & \textbf{10 Shot} & \textbf{15 Shot} \\ 
            \hline
            Ensemble 5 ResNets 4 DenseNets & 900.01 & 47.01 & Dense & 37.89 ± 0.99 & 62.07 ± 0.98 & 72.33 ± 0.81 & 76.93 ± 0.66 \\ 
            \hline
            \multirow{3}{5cm}{Ensemble MaxViT, ConvNeXt, Swin Transformer (Ours)} & \multirow{3}{*}{500.3} & \multirow{3}{*}{22.98} & Dense & 40.17 ± 0.96 & 65.05 ± 0.60 & 74.11 ± 0.47 & 78.71 ± 0.43 \\ 
            \cline{4-8}
            & & & LSTM & 46.87 ± 1.16 & 83.03 ± 0.85 & 92.66 ± 0.52 & 95.27 ± 0.29 \\ 
            \cline{4-8}
            & & & Bi-LSTM & \textbf{\textcolor{blue}{49.08 ± 1.15}} & \textbf{\textcolor{blue}{89.85 ± 0.73}} & \textbf{\textcolor{blue}{97.08 ± 0.34}} & \textbf{\textcolor{blue}{97.8 ± 0.44}} \\ 
            \hline
        \end{tabular}
    }
    \label{Result1}
\end{table*}

Hence, we conducted additional experiments with smaller feature extractors in the same experimental setup setting-1, utilizing Dense as the classifier without fine-tuning the models. From Table \ref{Result2}, we observe that the top four performing models are
Tiny\_Vit, MobilenetV2, MobileNetV3\_Large, and MobileNetV3\_Small. Although the performance of Tiny\_Vit was the best, we prioritized reducing memory requirements and enhancing computational efficiency over accuracy, given that the file sizes of MobilenetV2, MobileNetV3\_Large, and MobileNetV3\_small are significantly lower. The file size of Tiny\_Vit, at 81 MB, is twice that of the combined file sizes of MobilenetV2, MobileNetV3\_Large, and MobileNetV3\_small, which total 40.37 MB. Consequently, for further experimentation, we selected the ensemble of MobilenetV2, MobileNetV3\_Large, and MobileNetV3\_small.

\begin{table}[htbp]
    \centering
    \caption{Experiment with different lightweight models for feature extraction with Dense classifier}
    \label{Result2}
    \setlength{\tabcolsep}{3pt} 
    \resizebox{\linewidth}{!}{
     \renewcommand{\arraystretch}{3}
    \begin{tabular}{|l|c|c|c|c|c|c|}
        \hline
        \textbf{Model} & \textbf{Model Size (MB)} & \textbf{1 shot} & \textbf{5 shot} & \textbf{10 shot} & \textbf{15 shot} & \textbf{80 shot} \\
        \hline
        Tiny\_ViT & 81 & 38.86$\pm$1.04 & 67.64$\pm$0.69 & 78.28$\pm$0.49 & 82.31$\pm$0.34 & 89.93$\pm$0.27 \\
        \hline
        Mobilevit & 4.85 & 19.42$\pm$0.65 & 28.77$\pm$0.50 & 33.25$\pm$0.50 & 35.83$\pm$0.50 & 85.60$\pm$1.07 \\
        \hline
        \textcolor{blue}{MobileNetV3\_Large} & \textcolor{blue}{20.92} & \textcolor{blue}{34.48$\pm$0.92} & \textcolor{blue}{60.71$\pm$0.59} & \textcolor{blue}{70.97$\pm$0.57} & \textcolor{blue}{76.02$\pm$0.43} & \textcolor{blue}{88.86$\pm$0.25} \\
        \hline
        RegNetX & 10.24 & 34.47$\pm$0.87 & 58.64$\pm$0.65 & 67.56$\pm$0.54 & 72.46$\pm$0.41 & 84.93$\pm$0.28 \\
        \hline
       FastVit & 28.82 & 27.07$\pm$0.73 & 50.77$\pm$0.58 & 61.91$\pm$0.47 & 68.00$\pm$0.46 & 83.89$\pm$0.30 \\
        \hline
        \textcolor{blue}{MobileNetV3\_Small} & \textcolor{blue}{6.08} & \textcolor{blue}{36.99$\pm$0.91} & \textcolor{blue}{61.86$\pm$0.57} & \textcolor{blue}{71.37$\pm$0.51} & \textcolor{blue}{75.78$\pm$0.48} & \textcolor{blue}{87.66$\pm$0.28} \\
        \hline
        RegNetY & 16.80 & 28.56$\pm$0.74 & 50.80$\pm$0.61 & 59.93$\pm$0.47 & 64.76$\pm$0.55 & 80.18$\pm$0.29 \\
        \hline
        EfficientnetV0 & 20.17 & 35.26$\pm$0.87 & 58.30$\pm$0.61 & 66.00$\pm$0.50 & 74.40$\pm$0.44 & 84.23$\pm$0.27 \\
        \hline
        \textcolor{blue}{MobileNetV2} & \textcolor{blue}{13.37} & \textcolor{blue}{37.11$\pm$0.74} & \textcolor{blue}{62.64$\pm$0.60} & \textcolor{blue}{71.82$\pm$0.49} & \textcolor{blue}{76.07$\pm$0.42} & \textcolor{blue}{87.73$\pm$0.28} \\
        \hline
    \end{tabular}}
\end{table}

From Table \ref{Result3}, we observe that although the top-1 and top-5 accuracy of MaxViT, ConvNeXt, and Swin Transformer is higher than that of MobilenetV2, MobileNetV3\_Large, and MobileNetV3\_small in IMAGENET1K\_V1, the comparison of parameters, file size, and computational power indicates that MobilenetV2, MobileNetV3\_Large, and MobileNetV3\_small require significantly less memory and are computationally more efficient than MaxViT, ConvNeXt, and Swin Transformer.

\begin{table}[htbp]
    \centering
    \caption{Attributes of Our Previously Selected Feature Extractors Vs Current Feature Extractors}
    \label{Result3}
    \renewcommand{\arraystretch}{1.5} 
    \setlength{\tabcolsep}{3pt} 
    \resizebox{\linewidth}{!}{ 
    \begin{tabular}{|l|c|c|c|c|c|}
        \hline
        \textbf{Model Name} & \textbf{Parameters (M)} $\downarrow$ & \textbf{acc@1} $\uparrow$ & \textbf{acc@5} $\uparrow$ & \textbf{Size (Mb)} $\downarrow$ & \textbf{GFLOPs} $\downarrow$ \\
        \hline
        MAXVIT\_T & 30.9 & \textcolor{blue}{83.7} & \textcolor{blue}{96.722} & 118.8 & 5.56 \\
        \hline
        CONVNEXT\_SMALL & 50.2 & 83.61 & 96.65 & 191.7 & 8.68 \\
        \hline
        SWIN\_S & 49.6 & 83.196 & 96.36 & 189.8 & 8.74 \\
        \hline
        MobileNet\_V3\_Small & \textcolor{blue}{2.54} & 67.668 & 87.402 & \textcolor{blue}{9.8} & \textcolor{blue}{0.06} \\
        \hline
        MobileNet\_V3\_Large & 5.48 & 74.042 & 91.34 & 21.1 & 0.22 \\
        \hline
        MobileNet\_V2 & 3.5 & 71.87 & 90.28 & 13.6 & 0.3 \\
        \hline
    \end{tabular}%
    }
\end{table}

\subsubsection{Enhanced Performance Using Ensemble of Three Lightweight MobileNet Feature Extractors}
After selecting the ensemble of MobileNetV2, MobileNetV3\_small, and MobileNetV3\_large as our feature extractor, we pretrained each of these models on the PlantVillage dataset, excluding the tomato classes. We then used the pretrained weights in our experimental setup-1 and thoroughly analyzed the performance of this ensemble across various k-shot scenarios, including 1-shot, 5-shot, 10-shot, and 15-shot, using Dense, LSTM, and Bi-LSTM classifiers. As shown in Table \ref{Result4}, our framework, which uses the ensemble of MobileNetV2, MobileNetV3\_small, and MobileNetV3\_large as the feature extractor and Bi-LSTM as the classifier, outperformed our previous framework that used the ensemble of MAXVIT\_T, CONVNEXT\_S, and SWIN\_S as the feature extractor and Bi-LSTM as the classifier.

The reason for this superior performance could be that an ensemble of MobileNetV2, MobileNetV3\_large, and MobileNetV3\_small tends to perform better in few-shot learning scenarios with limited data compared to an ensemble of MAXVIT\_T, CONVNEXT\_S, and SWIN\_S. This is primarily due to the MobileNet models' design for efficiency and lower computational requirements. Table 4 shows that the total model size and GFlops of the ensemble of MobileNetV2, MobileNetV3\_small, and MobileNetV3\_large are significantly less than those of the ensemble of MAXVIT\_T, CONVNEXT\_S, and SWIN\_S. In addition, MobileNet models have fewer parameters and streamlined architectures, reducing the risk of overfitting on small data sets. They are optimized for efficiently extracting meaningful features, even with minimal data, and their pretrained weights transfer well to new tasks. In contrast, the larger and more complex architectures of MAXVIT, CONVNEXT, and SWIN often require more data to leverage their full potential and are more prone to overfitting when data is scarce. This is why the computationally efficient ensemble of MobileNetV2, MobileNetV3\_small, and MobileNetV3\_large as feature extractors gave better performance than the ensemble of MAXVIT\_T, CONVNEXT\_S, and SWIN\_S.

\begin{table*}[htbp] 
\centering
\caption{Comparative Performance of Lightweight MobileNet Ensemble versus Advanced Feature Extractor Ensemble in Experimental Setup-1}
\renewcommand{\arraystretch}{1.5} 
\setlength{\tabcolsep}{3pt} 
\resizebox{\linewidth}{!}{ 
\begin{tabular}{|p{5cm}|p{2cm}|c|c|c|c|c|c|}
\hline
\textbf{Feature Extractor} & \textbf{Total Model Size (MB)} & \textbf{GFLOPS} & \textbf{Classifier} & \textbf{1 Shot} & \textbf{5 Shot} & \textbf{10 Shot} & \textbf{15 Shot} \\ \hline
Ensemble MAXVIT\_T, CONVNEXT\_S, SWIN\_S & 500.3 & 22.98 & Bi-LSTM & 49.08 ± 1.15 & \textbf{\textcolor{blue}{89.85 ± 0.73}} & 97.08 ± 0.34 & 97.8 ± 0.44 \\ \hline
\multirow{3}{5cm}{Ensemble MobilenetV2, Mobilenetv3\_small, Mobilenetv3\_large} & \multirow{3}{*}{40.37} & \multirow{3}{*}{1.12} & Dense & 43.96 ± 1.0 & 70.46 ± 0.57 & 79.27 ± 0.45 & 82.76 ± 0.37 \\ \cline{4-8} 
 &  &  & LSTM & 51.66 ± 1.2 & 84.04 ± 0.92 & 91.83 ± 0.67 & 95.31 ± 0.42 \\ \cline{4-8} 
 &  &  & Bi-LSTM & \textbf{\textcolor{blue}{55.91 ± 1.34}} & 87.8 ± 1.3 & \textbf{\textcolor{blue}{98.44 ± 0.21}} & \textbf{\textcolor{blue}{98.44 ± 0.21}} \\ \hline
\end{tabular}%
}
\label{Result4}
\end{table*}

\subsubsection{Performance Incorporating Attention Mechanism with Classifier}

From Paper \cite{lina2021positive}, we learned that incorporating attention mechanisms with classifiers can significantly enhance plant disease classification. Attention mechanisms enable models to focus on specific regions of an image that are most relevant for making accurate predictions, effectively filtering out irrelevant information. In the context of plant disease classification, where subtle visual cues and patterns are crucial for accurate diagnosis, attention mechanisms enable the model to prioritize informative features while disregarding background noise or irrelevant areas.

That is why, using Experimental Setup Setting-1, we experimented with our framework using an ensemble of MobilenetV2, Mobilenetv3\_small, and Mobilenetv3\_large as feature extractors with different variations of classifiers incorporating attention mechanisms. We first tried only Self-Attention and then consecutively experimented with Bi-LSTM+Self-Attention and Bi-LSTM+Multihead Attention. We found that, according to Table \ref{tab:Result5}, the latter two performed better than only Self-Attention, but their results are quite close to using only Bi-LSTM as a classifier. This is because in PlantVillage, there is not enough variability in the background for attention to focus on a specific portion of the image, and Bi-LSTM already captures both-way dependency in sequence data. Furthermore, across all shot scenarios, Bi-LSTM with Self-Attention performed better than Bi-LSTM with Multihead Attention because the amount of data is less, and the power of Multihead Attention to capture local context is not as relevant.
\begin{table*}[htbp]
    \centering
    \caption{Performance of Ensemble: MobilenetV2, MobilenetV3\_Small, MobilenetV3\_Large Incorporating Attention Mechanism in Classifier in Experimental Setup-1}
    \label{tab:Result5}
    \setlength{\tabcolsep}{3pt} 
    \resizebox{\linewidth}{!}{%
        \renewcommand{\arraystretch}{2} 
        \footnotesize 
        \begin{tabular}{|p{3cm}|l|l|l|l|l|}
            \hline
            \textbf{Feature Extractor} & \textbf{Classifier} & \textbf{1 Shot} & \textbf{5 Shot} & \textbf{10 Shot} & \textbf{15 Shot} \\ 
            \hline
            \multirow{3}{3cm}{Ensemble of MobilenetV2, MobilenetV3\_small, MobilenetV3\_large (Ours)} & Self-Attention & 46.09 $\pm$ 1.03 & 72.19 $\pm$ 0.71 & 96.04 $\pm$ 0.59 & 83.77 $\pm$ 0.35 \\ 
            \cline{2-6} 
            & Bi-LSTM + Self-Attention & \textbf{\textcolor{blue}{56.91 $\pm$ 1.32}} & \textbf{\textcolor{blue}{87.18 $\pm$ 1.57}} & \textbf{\textcolor{blue}{95.45 $\pm$ 0.62}} & \textbf{\textcolor{blue}{98.23 $\pm$ 0.33}} \\ 
            \cline{2-6} 
            & Bi-LSTM + Multihead attention & 56.46 $\pm$ 1.43 & 85.1 $\pm$ 1.88 & 89.22 $\pm$ 2.55 & 89.78 $\pm$ 2.0 \\ 
            \hline
        \end{tabular}%
    }
\end{table*}

\subsection{Experiment With Apple, Blueberry and Cherry Leaf Class}
To evaluate the robustness of our implemented framework we compared its performance with other similar approaches by following Experimental Setup setting-2\\

\begin{table*}[htbp]
\centering
\caption{Performance Comparison on PlantVillage Dataset For 6 leaf diseases of apple, blueberry, and cherry plants in Experimental Setup-2}
\label{cherry}
 \renewcommand{\arraystretch}{1.5} 
\setlength{\tabcolsep}{3pt} 
\resizebox{\linewidth}{!}{
\begin{tabular}{|l|p{3.5cm}|p{3.5cm}|c|c|c|c|}
\hline
\textbf{Approach} & \textbf{Feature Extractor} & \textbf{Classifier} & \textbf{1 Shot} & \textbf{10 Shot} & \textbf{15 Shot} & \textbf{80 Shot} \\ \hline
Arg$\ddot{u}$eso\textit{ et al}. \cite{argueso2020fewshot} & \textbf{x} & Inception V3 Network & 55.5 & 77.0 & 80.0 & 90.0 \\ \hline
Wang \textit{et al}. \cite{wang2021imal} & \textbf{x} &  & 63.8 & \textbf{x} & 91.3 & 96.0 \\ \hline
\multirow{3}{*}{Ours} & \multirow{3}{3.5cm}{Ensemble of MobilenetV2, MobilenetV3\_small, MobilenetV3\_large} & Bi-LSTM & 78.93 $\pm$ 3.81 & 99.7 $\pm$ 0.15 & \textbf{{\textcolor{blue}{99.72 $\pm$ 0.12}}} & 99.68 $\pm$ 0.12 \\ \cline{3-7} 
 &  & Bi-LSTM + Self attention & \textbf{{\textcolor{blue}{78.92 $\pm$ 3.12}}} & \textbf{{\textcolor{blue}{99.7 $\pm$ 0.19}}} & 99.62 $\pm$ 0.12 & 99.68 $\pm$ 0.16 \\ \cline{3-7} 
 &  & Bi-LSTM + Multi-head Attention & 76.75 $\pm$ 3.75 & 99.7 $\pm$ 0.18 & 99.65 $\pm$ 0.25 & \textbf{{\textcolor{blue}{99.73 $\pm$ 0.13}}} \\ \hline
\end{tabular}
}
\end{table*}
Previously we have shown that in our implemented framework, with the ensemble of  MaxViT, ConvNeXt, and Swin Transformer as feature extractors, we employed three types of classifiers, namely Dense, LSTM, and Bi-LSTM, for the classification task, feeding them the combined feature vectors generated by the three extractors. We conducted the same experiment across different shot scenarios, including 1 shot, 10 shots, 15 shots, and 80 shots. It was evident that our framework with Bi-LSTM outperformed the results obtained by both Argueso \textit{et al}. \cite{argueso2020fewshot} and Wang \textit{et al}. \cite{wang2021imal} across all the shots.

Now, with the same experimental setup of settings-2, we experimented with our latest framework using an ensemble of MobilenetV2, Mobilenetv3\_small, and the Mobilenetv3\_large with classifiers Bi-LSTM, Bi-LSTM+Self Attention, and Bi-LSTM+Multihead Attention. From Table \ref{cherry}, it is evident that our framework even better outperforms the results of Argueso \textit{et al}. \cite{argueso2020fewshot} and Wang \textit{et al}. \cite{wang2021imal} across all the shots. in the same experimental setup, proving the robustness of our framework further.

\subsection{Experiment With Real Life Field Images }
\begin{table*}[htbp]
\centering
\caption{Performance on Dhan-Shomadhan Dataset For 5 leaf diseases of rice using field images in Experimental Setup-3}
\label{tab:real}
\setlength{\tabcolsep}{3pt} 
\resizebox{\linewidth}{!}{%
\renewcommand{\arraystretch}{2.5}
\begin{tabular}{|p{4cm}|l|l|l|l|l|}
\hline
\textbf{Feature Extractor} & \textbf{Classifier}        & \textbf{1 Shot} & \textbf{5 Shot} & \textbf{10 Shot} & \textbf{15 Shot} \\ 
\hline
\multirow{3}{4cm}{Ensemble of MobilenetV2, Mobilenetv3\_small, Mobilenetv3\_large} & Bi-LSTM & 32.88 $\pm$ 2.14  & 51.62 $\pm$ 2.06  & 59.64 $\pm$ 1.64  & 63.97 $\pm$ 1.69  \\ 
\cline{2-6} 
 & Bi-LSTM + Self-Attention & \textbf{{\textcolor{blue}{33.32 $\pm$ 2.41}}}  & 54.23 $\pm$ 1.83  & 60.97 $\pm$ 1.92  & 68.2 $\pm$ 1.82  \\ \cline{2-6} 
 & Bi-LSTM + Multihead Attention  & 33.01 $\pm$ 2.31  & \textbf{{\textcolor{blue}{55.68 $\pm$ 2.18}}}  & \textbf{{\textcolor{blue}{63.97 $\pm$ 1.66}}}  & \textbf{{\textcolor{blue}{69.28 $\pm$ 1.49}}}   \\ \hline
\end{tabular}%
}
\end{table*}
To evaluate the effectiveness of the attention mechanism incorporated with Bi-LSTM and to observe our framework's performance on real-field images, we conducted experiments on the Dhan Shomadhan \cite{hossain2023dhan} dataset. The results demonstrate the applicability of the attention mechanism in field images, consistently outperforming Bi-LSTM in each instance.
\newline
Our experimental setup includes a meta-training phase similar to other configurations, fine-tuning the feature extractors with 28 classes from the Plant Village Dataset to obtain a domain-adapted feature extractor. During the meta-testing phase, we utilized 5 classes of paddy leaf images, dividing the support and query sets in an 80:20 ratio, with the query set fully per class. We conducted 100 experiments for each shot and recorded the average performance.
\newline
The results in Table \ref{tab:real} indicate that both Self-Attention Bi-LSTM and Multihead Attention Bi-LSTM perform better than standalone Bi-LSTM for each k-shot scenario. This improvement is attributed to the attention mechanism's ability to focus on the actual leaf while ignoring the background, by considering contextual, local, and global features.
\newline

In laboratory images, where there is no background noise and the focus is solely on the leaf, the attention mechanism does not significantly enhance accuracy. Although the attention mechanism incorporated with Bi-LSTM produced results close to those of standalone Bi-LSTM, it ultimately outperformed Bi-LSTM in all instances.

\section*{Acknowledgment}

\bibliographystyle{IEEEtran}
\bibliography{citations.bib}

\end{document}